\documentclass[conference]{IEEEtran}
\IEEEoverridecommandlockouts

\usepackage{cite}
\usepackage{graphicx}
\usepackage{amsmath,amssymb}
\usepackage{booktabs}
\usepackage{hyperref}
\usepackage{siunitx}
\usepackage{microtype}
\usepackage{caption}
\usepackage{pgfplots}
\pgfplotsset{compat=1.18}
\usepackage{colortbl}
\usepackage{xcolor}
\usepackage{tabularx}
\definecolor{Gray}{gray}{0.9}

\begin{document}

\title{RodEpil: A Video Dataset of Laboratory Rodents for Seizure Detection and Benchmark Evaluation}

\author{
Daniele Perlo\textsuperscript{1},
Vladimir Despotovic\textsuperscript{2},
Selma Boudissa\textsuperscript{3},
Sang-Yoon Kim\textsuperscript{2}, \\
Petr V. Nazarov\textsuperscript{2,4},
Yanrong Zhang\textsuperscript{5},
Max Wintermark\textsuperscript{5},
Olivier Keunen\textsuperscript{1} \\ \\
\textsuperscript{1}\,BraiNE Unit, Department of Cancer Research, Luxembourg Institute of Health \\
\textsuperscript{2}\,Bioinformatics and AI Unit, Department of Medical Informatics, Luxembourg Institute of Health \\
\textsuperscript{3}\,Independent Researcher \\
\textsuperscript{4}\,Multiomics Data Science Research Group, Department of Cancer Research, Luxembourg Institute of Health \\
\textsuperscript{5}\,Department of Radiology, Stanford University
}

\maketitle

\begin{abstract}
We introduce a curated video dataset of laboratory rodents for automatic detection of convulsive events. The dataset contains short (10~s) top-down and side-view video clips of individual rodents, labeled at clip level as \emph{normal activity} or \emph{seizure}. It includes 10,101 negative samples and 2,952 positive samples collected from 19 subjects. We describe the data curation, annotation protocol and preprocessing pipeline, and report baseline experiments using a transformer-based video classifier (TimeSformer). Experiments employ five-fold cross-validation with strict subject-wise partitioning to prevent data leakage (no subject appears in more than one fold). Results show that the TimeSformer architecture enables discrimination between seizure and normal activity with an average F1-score of 97\%. The dataset and baseline code are publicly released to support reproducible research on non-invasive, video-based monitoring in preclinical epilepsy research. RodEpil Dataset access - DOI: 10.5281/zenodo.17601357
\end{abstract}

\begin{IEEEkeywords}
video dataset, seizure detection, rodents, TimeSformer, video classification, subject-wise cross-validation
\end{IEEEkeywords}

\section{Introduction}
Preclinical models of epilepsy commonly rely on electrophysiological monitoring using electroencephalography (EEG) and manual video scoring to quantify convulsive events. However, manual review of long recordings is labour intensive and limits throughput, while EEG monitoring is invasive and often impractical in large drug-screening studies. Video-based automatic seizure detection offers a non-invasive, scalable alternative that can accelerate behavioral phenotyping and
preclinical screening.
Recent advances in video representation learning — from 3D convolutional networks to transformer-based spatio-temporal models — suggests that visual features alone can be sufficiently discriminative for many motor seizure phenotypes \cite{bertasius2021timesformer,arnab2021vivit,feichtenhofer2019slowfast}.

To advance research in this direction, this paper introduces a labeled video dataset specifically collected for convulsive seizure classification in laboratory rodents, and provides baseline results obtained using a transformer-based TimeSformer model \cite{bertasius2021timesformer}. The dataset is intended to support the development and evaluation of non-invasive, video-based approaches for seizure detection in preclinical studies and to serve as a benchmark resource for future research in this domain. Our main contributions are as follows:

\begin{itemize}
\item A detailed description of the dataset structure, acquisition setup, and annotation protocol;
\item A reproducible experimental framework employing subject-wise five-fold cross-validation to ensure rigorous evaluation without data leakage;
\item Comprehensive baseline experiments using the TimeSformer architecture, including full preprocessing and training recipes; and
\item Release of the dataset and baseline code to enable transparent comparison and facilitate future research.
\end{itemize}

\section{Related work (State of the art)}
Video classification has evolved rapidly over the past decade. Early efforts used 3D convolutions (C3D) to learn spatio-temporal filters directly from video data \cite{tran2015c3d}. Inflated 3D (I3D) networks further improved the results by inflating successful 2D image architectures into 3D counterparts and pretraining on large-scale datasets such as Kinetics \cite{carreira2017quo, kay2017kinetics}. SlowFast networks introduced a dual-pathway design to separately model slow semantic information and fast motion \cite{feichtenhofer2019slowfast}. Recently, transformer-based methods (ViViT, TimeSformer) apply self-attention to spatio-temporal tokens and have shown state-of-the-art or competitive performance on standard benchmarks when properly regularized or pre-trained \cite{arnab2021vivit,bertasius2021timesformer}.

Building on these advances in video representation learning, several recent studies have applied both classical and deep learning methods to the analysis of convulsive events in preclinical models.
Approaches range from hand-crafted feature detectors and motion-based heuristics to deep convolutional and skeleton-based methods, and multi-modal combinations of EEG, motion sensors and video \cite{diaz2022videopackage, ren2024epidetect, mullen2024multimodal}. Reviews of video-based seizure analysis summarize advances in architectures and evaluation methodologies appropriate for clinical and preclinical settings \cite{ahmedt2024deeplearning_review}.

For tasks involving animal behavior, markerless pose estimation toolkits such as DeepLabCut and SLEAP have become standard for extracting fine-grained movement descriptors that can complement raw video classifiers \cite{mathis2018deeplabcut,pereira2020sleap}.
DeepLabCut uses deep convolutional networks (typically ResNet variants) to track individual body parts \cite{mathis2018deeplabcut}, while SLEAP employs encoder-decoder architectures such as UNet and can handle multiple animals simultaneously in top-down or bottom-up modes \cite{pereira2020sleap}. These CNN-based approaches excel at capturing spatial relationships between keypoints within individual frames but are limited in modeling long-range temporal dependencies across sequences. Because convulsive behaviors often unfold over extended time, capturing such temporal dynamics is essential. This motivates the use of transformer-based sequence models, which apply self-attention over spatio-temporal tokens and can learn complex patterns across extended time windows. In addition, unsupervised and anomaly detection approaches have also been proposed to identify abnormal behaviors in animal video streams \cite{nwokedi2021unsupervised}, which often involve unusual postures evolving over multiple frames, highlighting the need for models that integrate both spatial and temporal information.

This body of literature motivates the two key design choices in our baseline: (i) the use of a transformer-based model to capture long-range temporal dynamics, and (ii) strict subject-wise cross-validation to avoid optimistic bias when animals serve as the unit of replication \cite{kohavi1995cv}.

\section{Dataset}
\subsection{Acquisition and labelling}

The videos used in this study were collected in a controlled laboratory environment as part of an experimental study testing a non-invasive surgery approach for the treatment of epilepsy~\cite{zhang2021}. Epileptic rats were treated by systemically administered neurotoxin while the blood-brain-barrier was locally opened in a targeted region of the brain, using magnetic resonance guided focused ultrasounds in combination with intravenous microbubbles. 19 rodents with pilocarpine induced epilepsy were video recorded for 10 hours per day during 30 days before the treatment (from Jan 14 to Feb 19) and after the treatment (May 9 to June 8). Each video covered 3 to 6 cages. The frame size of the recordings were $640\times480$ pixels (VGA) and the frame rate 30 fps.

\begin{figure}[htbp]
\centerline{\includegraphics[scale=.5]{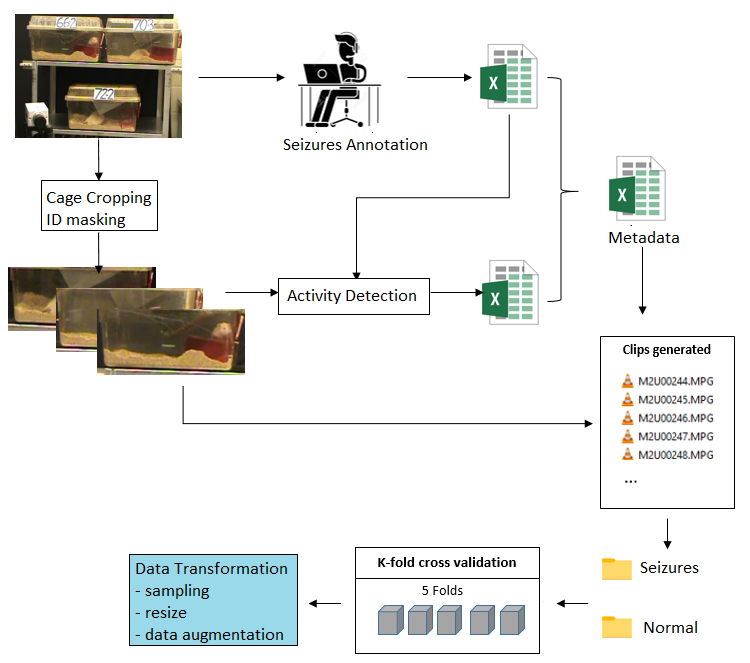}}
\caption{Video pre-processing pipeline}
\label{fig:data processing}
\end{figure}

Annotations of seizures were collected with the start and end time of each detected seizure  in the full videos recordings by trained operators by using standard operational criteria for convulsive motor seizures. Annotations were then peer-reviewed by a medical researcher with experience in experimental epilepsy and only clips with high inter-rater agreement were retained as labeled examples. With videos containing 3 to 6 cages by a fixed camera, single cages were first cropped from the full recording by manually drawing a bounding box around the cage on the first frame and applying it to the remaining frames. Bounding boxes were drawn to exclude the visible identification tag on cages to avoid unwanted biases, as illustrated in Fig.~\ref{fig:data processing}.

For each cage, the complete video sequences were further split into 10 second long clips that were assigned the positive class if a complete or partial seizure had been annotated for the corresponding period. From the clips that did not contain seizures, we only kept those in which motion could be detected by classical computer vision methods and excluded the clips in which the rats were simply laying still.

Filenames adhere to the following convention:
\vspace{0.3em}
\begin{quote}
\scalebox{0.8}{\texttt{Date\_VideoID\_Position\_RatID\_frames\_Label\_Y.avi}}
\end{quote}
\vspace{0.3em}
where \texttt{Label} is \texttt{N} for normal activity or \texttt{S} for seizure and \texttt{Y} is 0 (negative) or 1 (positive). 
For example, a sample filename is:
\vspace{0.3em}
\begin{quote}
\scalebox{0.8}{\texttt{Feb-1-2019\_M2U00244\_3\_70-3\_1153-1168\_N\_0.avi}}
\end{quote}
\vspace{0.3em}

\subsection{Statistics}
Table~\ref{tab:dataset} summarizes the composition of the final curated dataset. The dataset includes 19 individual subjects and consists of 10,101 negative clips showing normal activity and 2,952 positive clips containing convulsive events. Each clip is 10~seconds long, with frame rates standardized to 30~FPS during preprocessing. For training purposes, all frames were downsampled to a resolution of 224$\times$224 pixels. This composition ensures a substantial number of examples per class while maintaining uniform temporal and spatial characteristics across the dataset, which facilitates training and evaluation of video-based seizure classification models. The dataset exhibits a moderate class imbalance, with approximately three times as many negative clips as positive clips, which should be taken into account when designing and evaluating classification algorithms.

\begin{table}[ht]

\caption{Characteristics of the curated dataset.}
\label{tab:dataset}
\centering
\begin{tabularx}{5.5cm}{lc}
\toprule
Total subjects & 19 \\
Negative clips (normal activity) & 10,101 \\
Positive clips (convulsions) & 2,952 \\
Clip duration & 10~s \\
Frame rate & 30~FPS \\
Resolution & 224$\times$224 \\
\bottomrule
\end{tabularx}
\end{table}

\subsection{Data preprocessing}

Spatial resizing and center cropping were applied to produce input frames of 224$\times$224 pixels. Standard data augmentation, including random horizontal flips, random translations, small rotations, and jittering, was applied. For the baseline TimeSformer model, 16 frames were sampled uniformly per clip.

In our experiments, we evaluate three distinct input modalities, i.e., plain RGB frames, Absolute Differences, and Optical Flow, which capture complementary aspects of the video signal.
\begin{enumerate}
\item \textbf{RGB frames} represent standard video input consisting of raw pixel intensity values across the red, green, and blue channels. RGB frames preserve both spatial appearance (e.g., animal posture, background, and texture) and implicit temporal cues through frame ordering. It allows the model to jointly reason about visual context and motion over time, without explicitly computing motion features.
\item \textbf{Absolute Differences (AbsDiff)} encode motion by highlighting pixel-level intensity changes between consecutive frames. This approach emphasizes regions of movement while suppressing static background information, providing a simple and computationally efficient way to capture motion dynamics. However, it discards texture and appearance cues that may also carry behavioral information.
\item \textbf{Optical Flow} represents the apparent motion of pixels between consecutive frames, describing both the direction and magnitude of movement. Unlike Absolute Differences, it explicitly models motion trajectories, helping the model to recognize rhythmic or directional movement patterns characteristic of convulsive seizures.
\end{enumerate}

By comparing these modalities, we aim to disentangle the relative contribution of appearance information (captured by RGB) versus explicit motion cues (captured by Absolute Differences and Optical Flow).

\section{Methods}
\subsection{Model: TimeSformer}
TimeSformer operates on sequences of image patches and consists of 12 transformer blocks with an embedding dimension of 768. Spatial and temporal self-attention are factorised, reducing memory requirements while retaining expressive capacity \cite{bertasius2021timesformer}. For our experiments we use a standard TimeSformer backbone with the \textit{divided attention} variant, in which temporal attention is applied first, followed by spatial attention within each block. Where indicated, models are initialized from ImageNet-pretrained Vision Transformer (ViT) weights and further pre-trained on video Kinetics dataset.

\subsection{Implementation details}

All models were trained using 16 frame video clips sampled uniformly from each video. Each frame was divided into non-overlapping patches of size 16$\times$16 pixels, forming the input tokens for the transformer backbone.

Model weights were initialized from a Kinetics pretrained checkpoint to leverage large-scale video representation learning. Training was performed using stochastic gradient descent (SGD) with an initial learning rate of 0.005, weight decay of $1\times10^{-4}$, and a fixed step-decay learning rate schedule. Each GPU processed mini-batches of 8 video clips, and training was run for 20 epochs. To address class imbalance between seizure and non-seizure samples, a weighted cross-entropy loss was applied.

All experiments were conducted on a high-performance server equipped with eight NVIDIA RTX~A6000 graphics cards, each providing 48~GB of GPU memory. Model training was implemented in PyTorch, using distributed data parallelism to ensure efficient hardware utilization and scalability across multiple GPUs. Implementation followed best practices established in recent video-modeling literature, including gradient clipping for training stability, mixup augmentation when applicable, and deterministic seeding to ensure reproducibility.

\section{Experiments}
\subsection{Training regimes}
We evaluate three training regimes to examine the impact of pretraining and domain adaptation on seizure classification performance:
\begin{enumerate}
\item \textbf{Training from scratch}: The TimeSformer model is initialized from ImageNet-pretrained ViT weights and fine-tuned on the rodent video dataset.
\item \textbf{Kinetics pretraining + fine-tuning}: The TimeSformer model is first pre-trained on large-scale human action recognition datasets (Kinetics-400/600), followed by fine-tuning on the rodent dataset.
\end{enumerate}

\subsection{Evaluation}
For each regime, we employ 5-fold subject-wise cross-validation to ensure rigorous and unbiased evaluation. Each animal is assigned exactly to one fold, and all video clips from that subject are contained within the same fold. This protocol prevents information leakage arising from subject-specific motion or appearance patterns between training and test sets, providing realistic estimates of generalization to unseen animals.

Performance is evaluated using accuracy, precision (positive class), recall (sensitivity), and F1-score. Because the dataset is imbalanced, with substantially more normal activity clips than seizure events, precision and recall serve as more meaningful indicators of classification performance than accuracy alone, which can be inflated by the dominant class. The F1 score is the harmonic mean of precision and recall, providing a single metric that balances false positives and false negatives, making it a suitable evaluation metric for imbalanced datasets.

\section{Results}
We conducted a series of experiments to evaluate the TimeSformer model performance
under different training regimes and input data representations.
All models were trained for 20 epochs using 5-fold subject-wise cross-validation. 
Data augmentation was further used to expand the training dataset using flipping, translations, small rotations and jittering.

The performance of the model, pretrained on Kinetics-400/600 and fine-tuned on 16-frame RGB clips with the SGD optimizer, is detailed in Table~\ref{tab:best_model_results}. This configuration achieved an 
average accuracy of 98.55\% and an F1-score of 97.00\%.

\begin{table}[htbp]
    \centering
    \caption{Per-fold results for the best performing TimeSformer model (Kinetics pretraining+fine-tuning, RGB, 16 frames, batch size 8, SGD optimizer).}
    \label{tab:best_model_results}
    \begin{tabularx}{\columnwidth}{lXXXXXX}
    \toprule
    \textbf{CV} & \textbf{Accuracy} & \textbf{Sensitivity} & \textbf{Specificity} & \textbf{Precision} & \textbf{F1-score} \\
    \midrule
    fold 0 & 98.95\% & 99.67\% & 98.86\% & 91.82\% & 95.58\% \\
    fold 1 & 99.13\% & 100.00\% & 98.96\% & 95.04\% & 97.46\% \\
    fold 2 & 99.49\% & 99.87\% & 99.33\% & 98.44\% & 99.15\% \\
    fold 3 & 99.58\% & 98.96\% & 99.69\% & 98.35\% & 98.66\% \\
    fold 4 & 95.59\% & 90.56\% & 98.77\% & 97.90\% & 94.09\% \\
    \midrule
    \textbf{Average} & \textbf{98.55\%} & \textbf{97.81\%} & \textbf{99.12\%} & \textbf{96.31\%} & \textbf{97.00\%} \\
    \bottomrule
    \end{tabularx}
\end{table}

To understand the contribution of different components, we performed an ablation study, summarized in Table~\ref{tab:ablation_summary}. The results indicate that using RGB frames, transfer learning from the human motion via Kinetics pretraining, and the larger spatial resolution and temporal coverage, were critical for achieving peak performance.

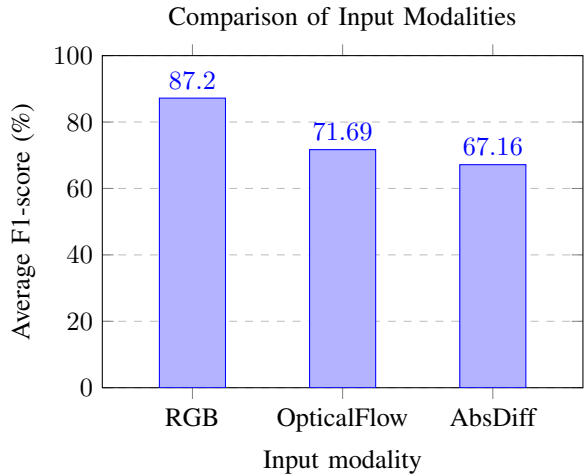
\begin{figure}[htbp]
    \centering
    \begin{tikzpicture}
        \begin{axis}[
            ybar,
            width=0.9\columnwidth,
            height=6cm,
            bar width=25pt,
            enlarge x limits=0.3,
            title={Comparison of Input Modalities},
            xlabel={Input modality},
            ylabel={Average F1-score (\%)},
            symbolic x coords={RGB, OpticalFlow, AbsDiff},
            xtick=data,
            nodes near coords,
            nodes near coords align={vertical},
            ymin=0, ymax=100,
            ymajorgrids=true,
            grid style=dashed,
        ]
        \addplot coordinates {(RGB, 87.20) (OpticalFlow, 71.69) (AbsDiff, 67.16)};
        \end{axis}
    \end{tikzpicture}
    \caption{Comparison of average F1-scores for models trained on RGB, Optical Flow, and Absolute Difference inputs (224x224, 8 frames, SGD, Kinetics pretrain).}
    \label{fig:input_comparison}
\end{figure}

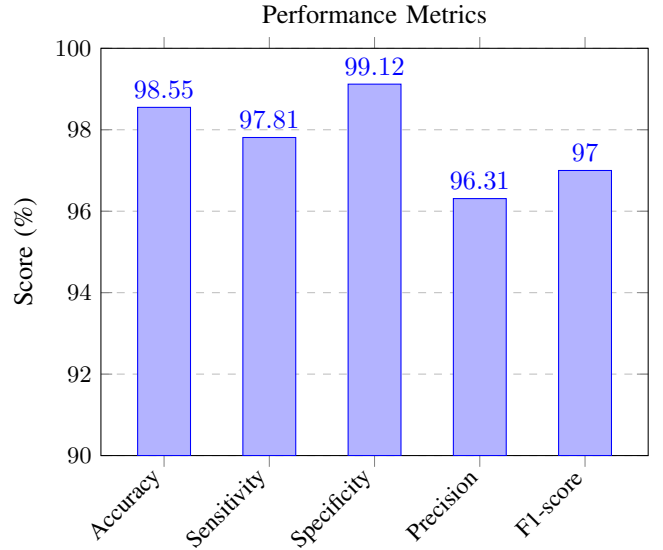
\begin{figure}[htbp]
    \centering
    \begin{tikzpicture}
        \begin{axis}[
            ybar,
            width=\columnwidth,
            height=7cm,
            bar width=20pt,
            enlarge x limits=0.15,
            title={Performance Metrics},
            ylabel={Score (\%)},
            symbolic x coords={Accuracy, Sensitivity, Specificity, Precision, F1-score},
            xtick=data,
            tick label style={font=\small},
            xticklabel style={rotate=45, anchor=east},
            nodes near coords,
            nodes near coords align={vertical},
            ymin=90, ymax=100,
            ymajorgrids=true,
            grid style=dashed,
        ]
        \addplot coordinates {
            (Accuracy, 98.55) 
            (Sensitivity, 97.81) 
            (Specificity, 99.12) 
            (Precision, 96.31) 
            (F1-score, 97.00)
        };
        \end{axis}
    \end{tikzpicture}
    \caption{Average performance metrics for the best performing model configuration across 5 folds.}
    \label{fig:best_model_metrics}
\end{figure}

Performance was further evaluated for different input video modalities, to analyze the influence of visual information versus explicit motion cues. As shown in Fig.~\ref{fig:input_comparison}, RGB inputs significantly outperform motion-based representations such as Optical Flow and Absolute Difference. Furthermore, Fig.~\ref{fig:best_model_metrics} provides a visual summary of the balanced and high-quality performance of our best model across multiple metrics.

\begin{table*}[htbp]
    \centering
    \caption{Summary of average performance across different experimental configurations.}
    \label{tab:ablation_summary}
    \begin{tabular}{ccccc|cccc}
    \toprule
    \textbf{Image Type} & \textbf{Img Size} & \textbf{\# Frames} & \textbf{Optim} & \textbf{Pretrained} & \textbf{Accuracy} & \textbf{Sensitivity} & \textbf{Precision} & \textbf{F1-score} \\
    \midrule
    \rowcolor{Gray}
    \textbf{RGB} & \textbf{224} & \textbf{16} & \textbf{SGD} & \textbf{Kinetics} & \textbf{98.55\%} & \textbf{97.81\%} & \textbf{96.31\%} & \textbf{97.00\%} \\
    RGB & 224 & 8 & SGD & Kinetics & 95.22\% & 85.02\% & 89.60\% & 87.20\% \\
    OpticalFlow & 224 & 8 & SGD & Kinetics & 88.01\% & 77.42\% & 68.32\% & 71.69\% \\
    AbsDiff & 224 & 8 & SGD & Kinetics & 85.84\% & 71.65\% & 64.91\% & 67.16\% \\
    RGB & 112 & 16 & SGD & Kinetics & 84.94\% & 67.70\% & 64.42\% & 64.39\% \\
    RGB & 112 & 8 & SGD & Kinetics & 82.59\% & 51.84\% & 68.34\% & 54.50\% \\
    RGB & 224 & 16 & AdamW & Kinetics & 74.61\% & 43.52\% & 28.68\% & 34.06\% \\
    RGB & 224 & 16 & SGD & no & 71.04\% & 48.18\% & 41.84\% & 38.57\% \\
    RGB & 224 & 16 & AdamW & no & 73.74\% & 24.74\% & 29.81\% & 25.61\% \\
    \bottomrule
    \end{tabular}
\end{table*}

\section{Discussion}
Transformer-based models such as TimeSformer 
are highly effective at capturing
long-range temporal dependencies and can disambiguate subtle motion patterns associated with convulsive events from normal grooming or locomotion. The subject-wise cross-validation protocol demonstrates excellent
generalization capability across animals and avoids inflated estimates caused by subject-overlap between training and test sets. 

Our baseline experiments, summarized in Table~\ref{tab:ablation_summary}, 
systematically evaluate the contribution of input modality, pretraining, image resolution, and sequence length. The results reveal several clear trends. Models trained with RGB frames consistently achieve the highest performance, reaching an average accuracy of 98.55\% and an F1-score of 97.00\%. This strong advantage over motion-based representations such as Optical Flow (88.01\% accuracy) and Absolute Differences (85.84\% accuracy) suggests that the contextual visual information, such as posture, environment, and lighting, plays an important role in discriminating seizure events beyond pure motion features. While motion-based inputs highlight temporal change, they remove static spatial details that are often discriminative for identifying specific postures or body configurations associated with seizures. In contrast, RGB inputs preserve both spatial structure and temporal evolution, enabling the model to learn richer and more generalizable representations.

Pretraining on the large-scale Kinetics dataset provides a substantial boost in performance. Models initialized with Kinetics-pretrained weights consistently outperform those trained from scratch by over 25\% in accuracy, highlighting the critical role of transfer learning in data-limited biomedical video domains. This improvement reflects the strong inductive bias that large-scale pretraining imparts, allowing the model to leverage general video dynamics before fine-tuning on a smaller, domain-specific dataset. The pretrained features appear to transfer well to rodent behavioral videos, despite the significant domain gap between natural human actions and laboratory animal movements.

Both spatial resolution and temporal coverage also play a measurable role. Reducing input resolution from 224$\times$224 to 112$\times$112 or halving the number of input frames from 16 to 8 leads to significant performance degradation. These results emphasize the importance of retaining fine-grained spatial detail and sufficient temporal context to capture convulsive patterns, which often involve localized limb movements evolving over several seconds. The superior performance of the 16-frame configuration suggests that the transformer benefits from broader temporal windows for modeling the onset and propagation of seizure-related activity.

Despite these encouraging results, several limitations remain. First, 
the dataset focuses on visible motor convulsions and 
does
not capture purely electrographic seizures without obvious motor manifestations.
Second, variability in lighting conditions, background contrast, and camera placement may affect generalization to other laboratory environments or recording setups.
Finally, the relatively short 10-second clips constrain the available temporal context and may limit the ability to detect early pre-convulsive behaviors or subtle prodromal signs.


\section{Conclusion}
We introduce in this paper
a curated dataset of short
rodent video clips annotated
for convulsive seizures and established a robust baseline using a TimeSformer model under
a strict 5-fold subject-wise cross-validation protocol. Our experiments demonstrate that visual-only classification of convulsive events is highly effective when sufficient spatial and temporal context is provided, achieving
an average accuracy of \textbf{98.55\%} and an F1-score of \textbf{97.00\%}. 

Performance was driven by several key factors. RGB inputs outperformed motion-based representations, emphasizing the importance of spatial detail and full visual context. Kinetics pretraining substantially improved generalization, underscoring the value of transfer learning in small-data regimes. These results highlight the sensitivity of transformer-based models to training configuration and the benefits of carefully tuned, pretrained architectures.

All dataset metadata, preprocessing scripts, and baseline code are made publicly available to encourage reproducible research and to provide a standardized benchmark for future work in automated behavioral phenotyping and video-based seizure detection.


Future work could extend these results by incorporating longer temporal windows for early detection, integrating complementary modalities such as EEG or piezo sensors to capture non-motor events, and combining raw video with pose-estimation outputs (e.g., DeepLabCut, SLEAP) to enhance interpretability. Expanding the dataset across diverse recording conditions and experimental protocols would further improve robustness and enable broader benchmarking of automated, non-invasive seizure monitoring in preclinical research.

\section*{Acknowledgments}
We thank the laboratory personnel of Stanford University who provided animal care and helped with video acquisition and annotation

\subsection{Ethical considerations and data sharing}
Data collection was carried out according to institutional animal-care guidelines. RodEpil Dataset access - DOI: 10.5281/zenodo.17601357.

\end{document}